%% file: deepmath.tex
\documentclass{article}
\usepackage[nonatbib,final,arxiv]{nips_2016}

\usepackage{nips_2016}

\usepackage[utf8]{inputenc} 
\usepackage[T1]{fontenc}    
\usepackage{hyperref}       
\usepackage{url}            
\usepackage{booktabs}       
\usepackage{amsfonts}       
\usepackage{nicefrac}       
\usepackage{microtype}      
\usepackage{color}
\usepackage{amsmath}
\usepackage{amsthm}
\usepackage{graphicx}
\usepackage{subcaption}
\usepackage{stackengine}
\usepackage{color}

\let\OLDthebibliography\thebibliography
\renewcommand\thebibliography[1]{
  \OLDthebibliography{#1}
  \setlength{\parskip}{2pt}
  \setlength{\itemsep}{2pt plus 0.3ex}
}

\newtheorem*{definition}{Definition}

\title{DeepMath - Deep Sequence Models for Premise Selection}

\author{
  Alexander A. Alemi \thanks{Authors listed alphabetically.  All contributions are considered equal.} \\
  Google Inc. \\
  \texttt{alemi@google.com} \\
  \And
  Fran\c{c}ois Chollet \footnotemark[1] \\
  Google Inc. \\
  \texttt{fchollet@google.com} \\
  \And
  Niklas Een \footnotemark[1] \\
  Google Inc. \\
  \texttt{een@google.com} \\
  \And
  Geoffrey Irving \footnotemark[1] \\
  Google Inc. \\
  \texttt{geoffreyi@google.com} \\
  \And
  Christian Szegedy \footnotemark[1] \\
  Google Inc. \\
  \texttt{szegedy@google.com} \\
  \And
  Josef Urban \footnotemark[1] \thanks{Supported by ERC Consolidator grant nr. 649043 \textit{AI4REASON}.}\\
  Czech Technical University in Prague \\
  \texttt{josef.urban@gmail.com} \\
}
\begin{document}

\maketitle

\begin{abstract}
  We study the effectiveness of neural sequence models for
  premise selection in automated theorem proving, one of the main bottlenecks in the formalization of
	mathematics.
   We propose a two stage approach for this task
  that yields good results for the premise selection task on the Mizar corpus while
  avoiding the hand-engineered features of existing state-of-the-art models.
  To our knowledge, this is the first time deep learning has been applied to
  theorem proving on a large scale.
\end{abstract}

\section{Introduction}

Mathematics underpins all scientific disciplines.  Machine learning itself
rests on measure and probability theory, calculus, linear algebra, functional
analysis, and information theory.  Complex mathematics underlies computer chips,
transit systems, communication systems, and financial infrastructure -- thus
the correctness of many of these systems can be reduced to mathematical proofs.

Unfortunately, these correctness proofs are often impractical to produce
without automation, and present-day computers have only limited ability to
assist humans in developing mathematical proofs and formally verifying human
proofs.  There are two main bottlenecks: (1) lack of automated methods for
\emph{semantic} or \emph{formal} parsing of informal mathematical texts
(\emph{autoformalization}), and (2) lack of strong automated reasoning methods
to fill in the gaps in already formalized human-written proofs.

The two bottlenecks are related. Strong automated reasoning can act as a
semantic filter for autoformalization, and successful autoformalization would
provide a large corpus of computer-understandable facts, proofs, and theory
developments.  Such a corpus would serve as both background knowledge to fill in
gaps in human-level proofs and as a training set to guide automated reasoning.
Such guidance is crucial: exhaustive deductive reasoning tools such as today's
resolution/superposition automated theorem provers (ATPs) quickly hit
combinatorial explosion, and are unusable when reasoning
with a very large number of facts without careful selection
\cite{BlanchetteKPU16}.

In this work, we focus on the latter bottleneck. We develop deep neural networks
that learn from a large repository of manually formalized
computer-understandable proofs.  We learn the task that is essential for making
today's ATPs usable over large formal corpora: the selection of a limited number of
most relevant facts for proving a new conjecture.  This is known as
\emph{premise selection}.

The main contributions of this work are:
\begin{itemize}
  \item A demonstration for the first time that neural network models are
        useful for aiding in large scale automated logical reasoning without
        the need for hand-engineered features.
  \item The comparison of various network architectures (including convolutional,
        recurrent and hybrid models) and their effect on premise selection
        performance.
  \item A method of semantic-aware ``definition''-embeddings for function
        symbols that improves  the generalization of formulas with symbols
        occurring infrequently. This model outperforms previous approaches.
  \item Analysis showing that  neural network based premise selection
        methods are complementary to those with hand-engineered features:
        ensembling with previous results produce superior results.
\end{itemize}

\section{Formalization and Theorem Proving}

In the last two decades, large corpora of complex mathematical
knowledge have been \emph{formalized}: encoded in complete detail so that
computers can \emph{fully understand the semantics} of complicated mathematical
objects. The process of writing such formal and
verifiable theorems, definitions, proofs, and theories is called
\emph{Interactive Theorem Proving} (ITP).

The ITP field dates back to 1960s~\cite{HarrisonUW14} and the Automath system
by N.G. de Bruijn~\cite{DeBruijn68}.  ITP systems include HOL
(Light)~\cite{Harrison96}, Isabelle~\cite{WenzelPN08},
Mizar~\cite{mizar-in-a-nutshell}, Coq~\cite{coq}, and ACL2~\cite{KaufmannM08}.
The development of ITP has been intertwined with the
development of its cousin field of \emph{Automated Theorem Proving}
(ATP)~\cite{DBLP:books/el/RobinsonV01},
where proofs of conjectures are
attempted fully automatically.  Unlike ATP systems, ITP systems allow
human-assisted formalization and proving of theorems that are often beyond the
capabilities of the fully automated systems.

Large ITP libraries
include the Mizar Mathematical Library (MML) with over 50,000 lemmas, and the
core Isabelle, HOL, Coq, and ACL2 libraries with thousands of lemmas.  These
core libraries are a basis for large projects in formalized mathematics and
software and hardware verification.  Examples in mathematics include the HOL
Light proof of the Kepler conjecture (Flyspeck
project)~\cite{HalesABDHHKMMNNNOPRSTTTUVZ15}, the Coq proofs of the
Feit-Thompson theorem~\cite{DBLP:conf/itp/GonthierAABCGRMOBPRSTT13} and Four
Color theorem~\cite{Gonthier07}, and the verification of most of the Compendium of Continuous Lattices in Mizar~\cite{BancerekR02}.
ITP verification of the seL4 kernel~\cite{KleinAEHCDEEKNSTW10} and CompCert
compiler~\cite{Leroy09} show comparable progress in large scale software verification.
While these large projects mark a coming of age of formalization, ITP remains
labor-intensive.  For example, Flyspeck took about 20 person-years,
with twice as much for Feit-Thompson. Behind this cost are our two bottlenecks:
lack of tools for autoformalization and strong proof automation.

Recently the field of
\emph{Automated Reasoning in Large Theories} (ARLT)~\cite{UrbanV13} has developed,
including AI/ATP/ITP (AITP) systems called \emph{hammers} that assist ITP
formalization~\cite{BlanchetteKPU16}.  Hammers analyze the full set of theorems
and proofs in the ITP libraries, estimate the relevance of each
theorem, and apply optimized translations from the ITP logic to simpler ATP
formalism.  Then they attack new conjectures using the most promising
combinations of existing theorems and ATP search strategies.  Recent
evaluations have proved 40\% of all Mizar and Flyspeck theorems fully
automatically~\cite{holyhammer,KaliszykU13b}.
However, there is significant room for improvement: with perfect premise selection
(a perfect choice of library facts) ATPs can prove at least 56\%
of Mizar and Flyspeck instead of today's 40\%~\cite{BlanchetteKPU16}.
In the next section we explain the premise selection task and the experimental
setting for measuring such improvements.

\section{Premise Selection, Experimental Setting and Previous Results \label{premise}}
\label{previous}
Given a formal corpus of facts and proofs expressed in an
ATP-compatible format, our task is
\begin{definition}[Premise selection problem]
Given a large set of premises $\mathcal{P}$, an ATP system $A$ with given
resource limits, and a new conjecture $C$, predict those premises from
$\mathcal{P}$ that will most likely lead to an automatically constructed
proof of $C$ by $A$.
\end{definition}

\begin{figure}
\begin{small}
\begin{verbatim}
:: t99_jordan: Jordan curve theorem in Mizar
for C being Simple_closed_curve holds C is Jordan;

:: Translation to first order logic
fof(t99_jordan, axiom,  (! [A] :  ( (v1_topreal2(A) & m1_subset_1(A,
k1_zfmisc_1(u1_struct_0(k15_euclid(2)))))  => v1_jordan1(A)) ) ).
\end{verbatim}
\end{small}
\caption{\small (top) The final statement of the Mizar formalization of the Jordan
curve theorem.  (bottom) The translation to first-order logic, using name
mangling to ensure uniqueness across the entire corpus. \label{fol}
}
\end{figure}

We use the Mizar Mathematical Library (MML)
version
4.181.1147\footnote{\url{ftp://mizar.uwb.edu.pl/pub/system/i386-linux/mizar-7.13.01_4.181.1147-i386-linux.tar}}
as the formal corpus and E prover~\cite{Sch02-AICOMM} version 1.9 as the underlying ATP system.
The following list exemplifies a small non-representative sample of topics and theorems that are included in the
Mizar Mathematical Library: Cauchy-Riemann Differential Equations of Complex Functions,
Characterization and Existence of {G}r\"obner Bases, Maximum Network Flow Algorithm by Ford and Fulkerson,
G{\"o}del's Completeness Theorem, Brouwer Fixed Point Theorem, Arrow's Impossibility Theorem
Borsuk-Ulam Theorem, Dickson's Lemma, Sylow Theorems, Hahn Banach Theorem, The Law of Quadratic Reciprocity,
Pepin's Primality Test for Public-Key Cryptography,  Ramsey's Theorem.


This version of MML was used for the latest AITP
evaluation reported in~\cite{KaliszykU13b}. There are 57,917 proved
Mizar theorems and unnamed top-level lemmas in this MML organized into
1,147 articles.  This set is chronologically ordered by
the order of articles in MML and by the order of
theorems in the articles. Proofs of later theorems can only refer to earlier theorems.
This ordering also applies to 
88,783 other Mizar formulas (encoding the type system and
other automation known to Mizar) used in the problems. The formulas have been translated into
first-order logic formulas by the MPTP system~\cite{Urban06} (see Figure~\ref{fol}). 

Our goal is to automatically prove as many theorems as
possible, using at each step all previous theorems and proofs. 
We can learn from both human proofs and ATP proofs, but previous
experiments~\cite{KuhlweinU12b,holyhammer} show that learning 
only from the 
ATP proofs is preferable to including human
proofs if the set of ATP proofs is sufficiently large.  Since for
32,524 (56.2\%) of the 57,917 theorems an ATP proof was previously
found by a combination of manual and learning-based premise
selection~\cite{KaliszykU13b}, we use only these ATP proofs for
training.

The 40\% success rate
from~\cite{KaliszykU13b} used a portfolio of 14 AITP methods
using different learners, ATPs, and numbers of premises. The best
single method proved 27.3\% of the theorems. Only fast and simple
learners such as $k$-nearest-neighbors, naive Bayes, and their ensembles were
used, based on hand-crafted features such as the set of
(normalized) sub-terms and symbols in each formula.

\input{deep}

\begin{small}

\bibliography{ate11}
\bibliographystyle{abbrv}

\end{small}

\end{document}

%% file: deep.tex

\section{Motivation for the use of Deep Learning}

Strong premise selection requires models capable of reasoning over
mathematical statements, here encoded
as variable-length strings of first-order logic.
In natural language processing, deep neural networks have
proven useful in language
modeling~\cite{mikolov2010recurrent},
text classification~\cite{dai2015semi},
sentence pair scoring~\cite{baudis2016sentence},
conversation modeling~\cite{vinyals2015neural}, and
question answering~\cite{sukhbaatar2015end}.
These results have demonstrated
the ability of deep networks to extract useful
representations from sequential inputs without
hand-tuned feature engineering. Neural networks can also
mimic some higher-level reasoning on simple algorithmic
tasks~\cite{
  zaremba2014learning, kaiser2015neural}.

\begin{figure}
\begin{subfigure}[t]{.24\textwidth}
 \centering
  \includegraphics[width=1.4in]{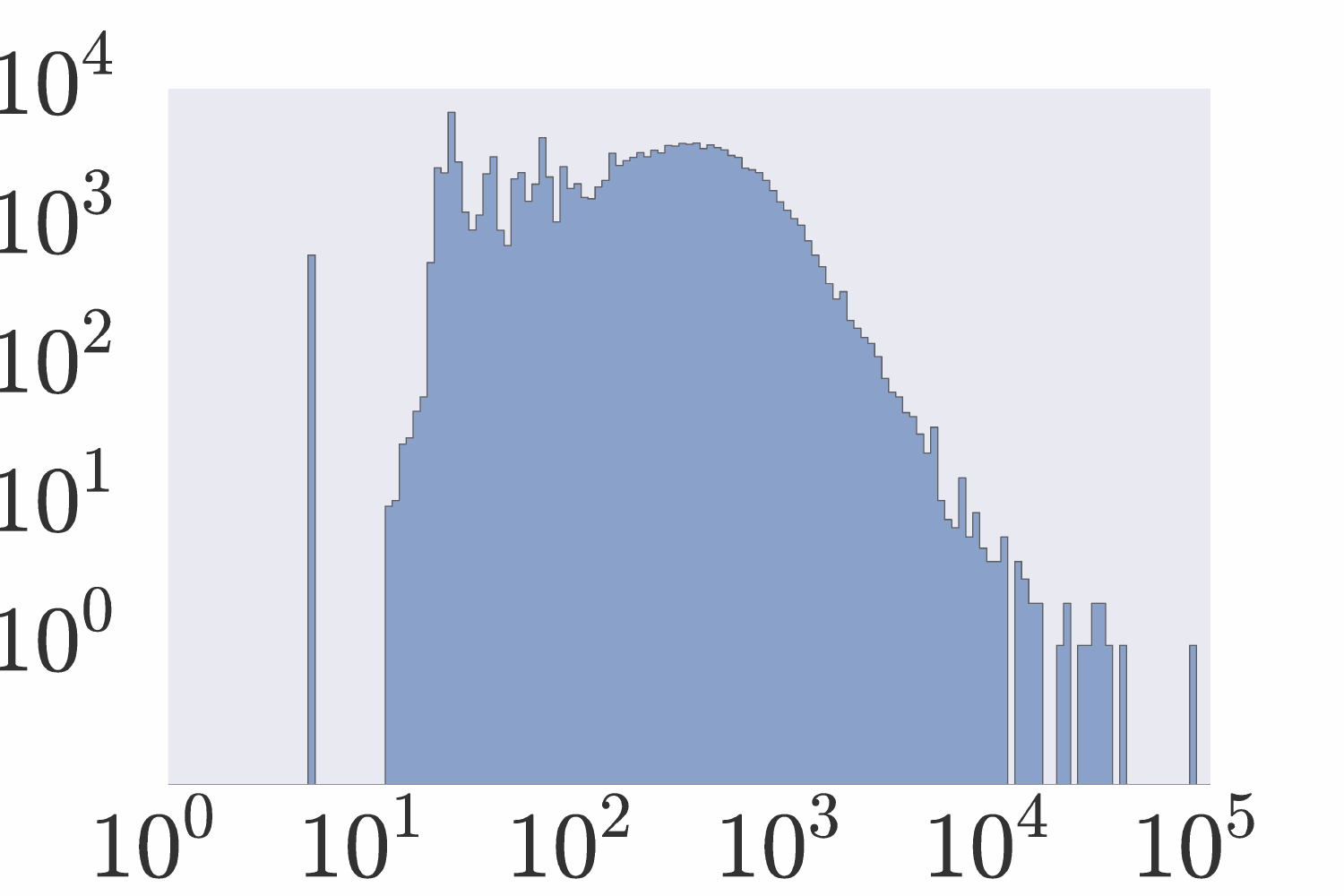}
  \caption{\small Length in chars.}
\end{subfigure}
\begin{subfigure}[t]{.24\textwidth}
  \centering
  \includegraphics[width=1.4in]{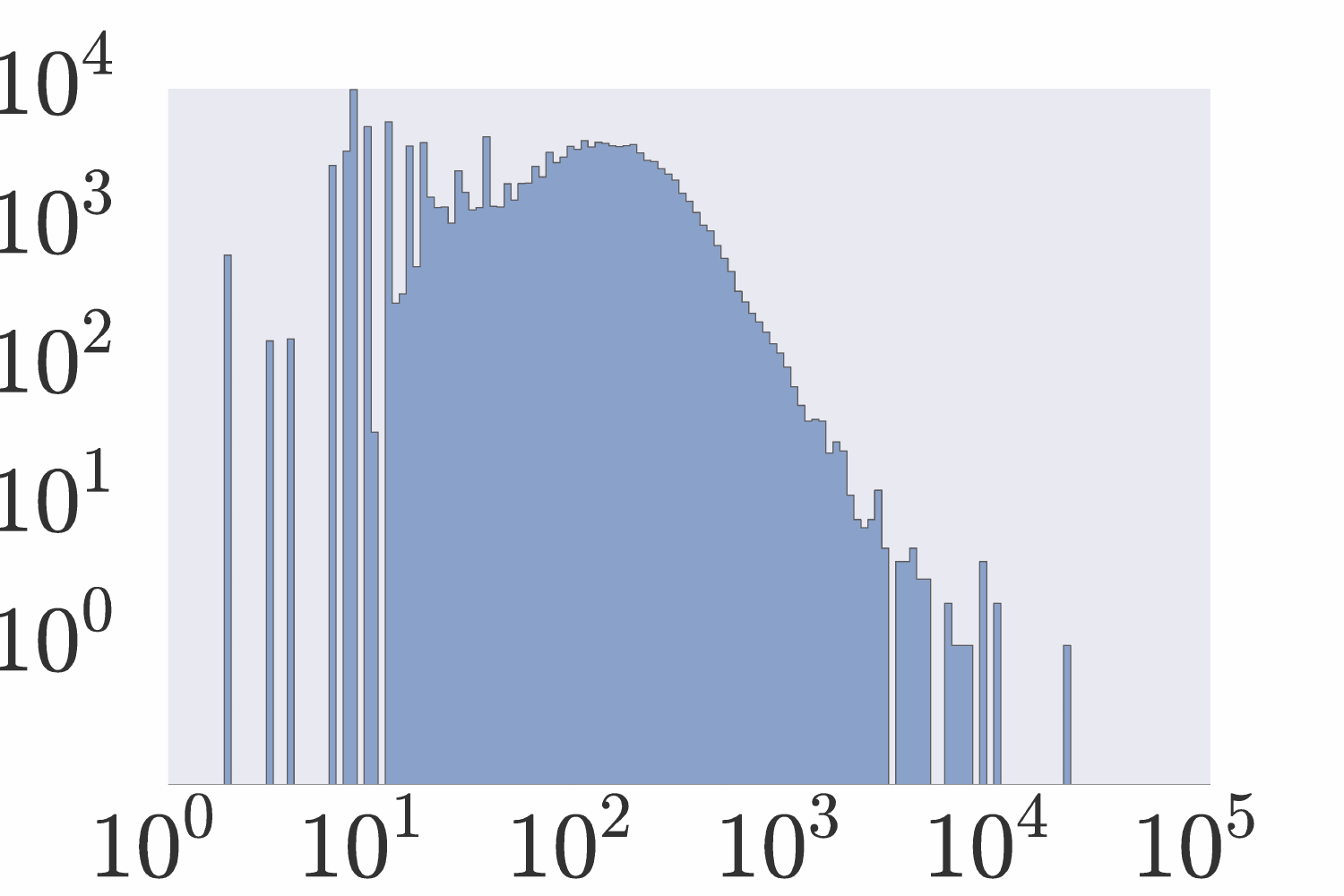}
  \caption{\small Length in words.}
\end{subfigure}
\begin{subfigure}[t]{.24\textwidth}
  \centering
  \includegraphics[width=1.4in]{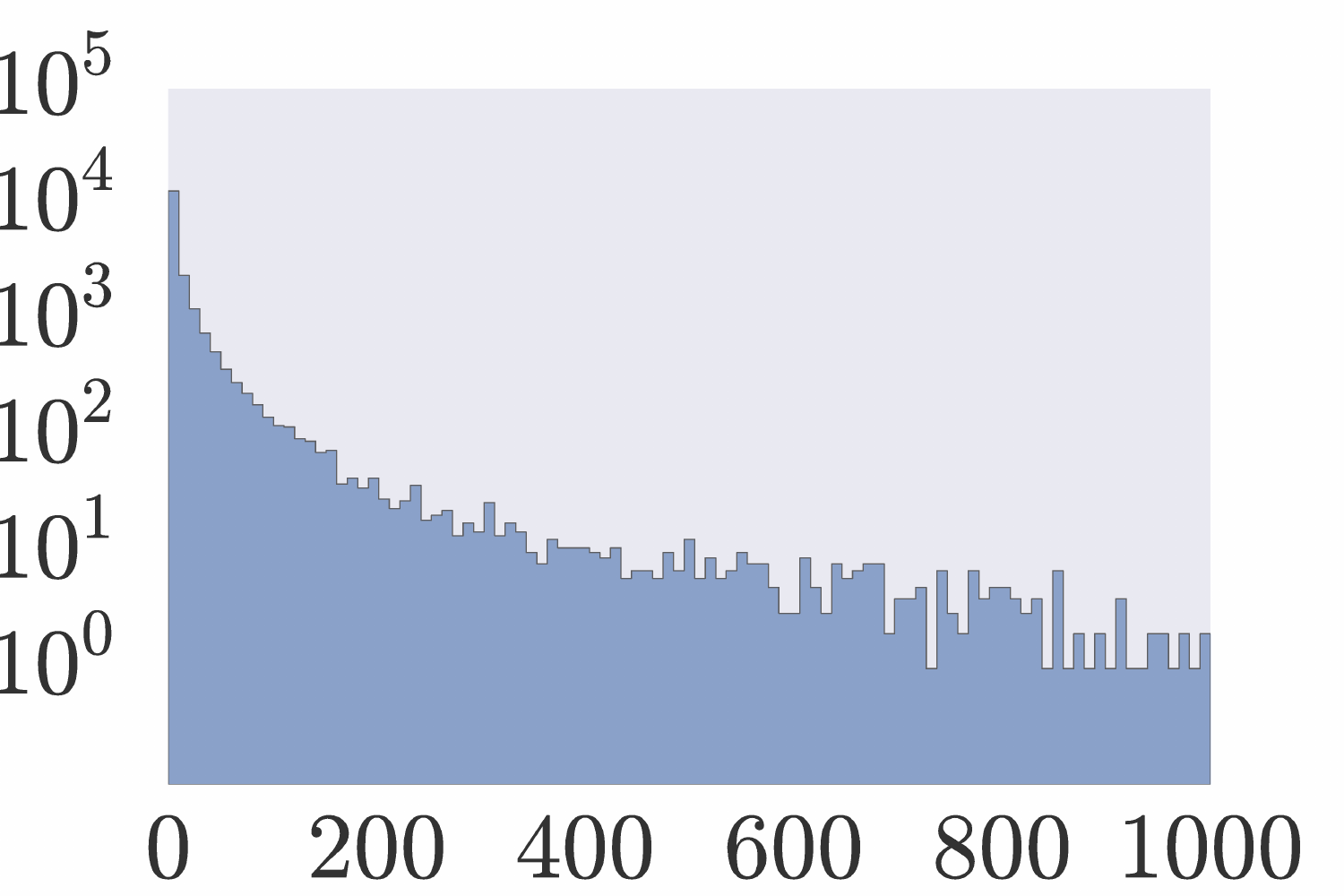}
  \caption{\small Word occurrences.}
\end{subfigure}
  \begin{subfigure}[t]{.24\textwidth}
  \centering
  \includegraphics[width=1.4in]{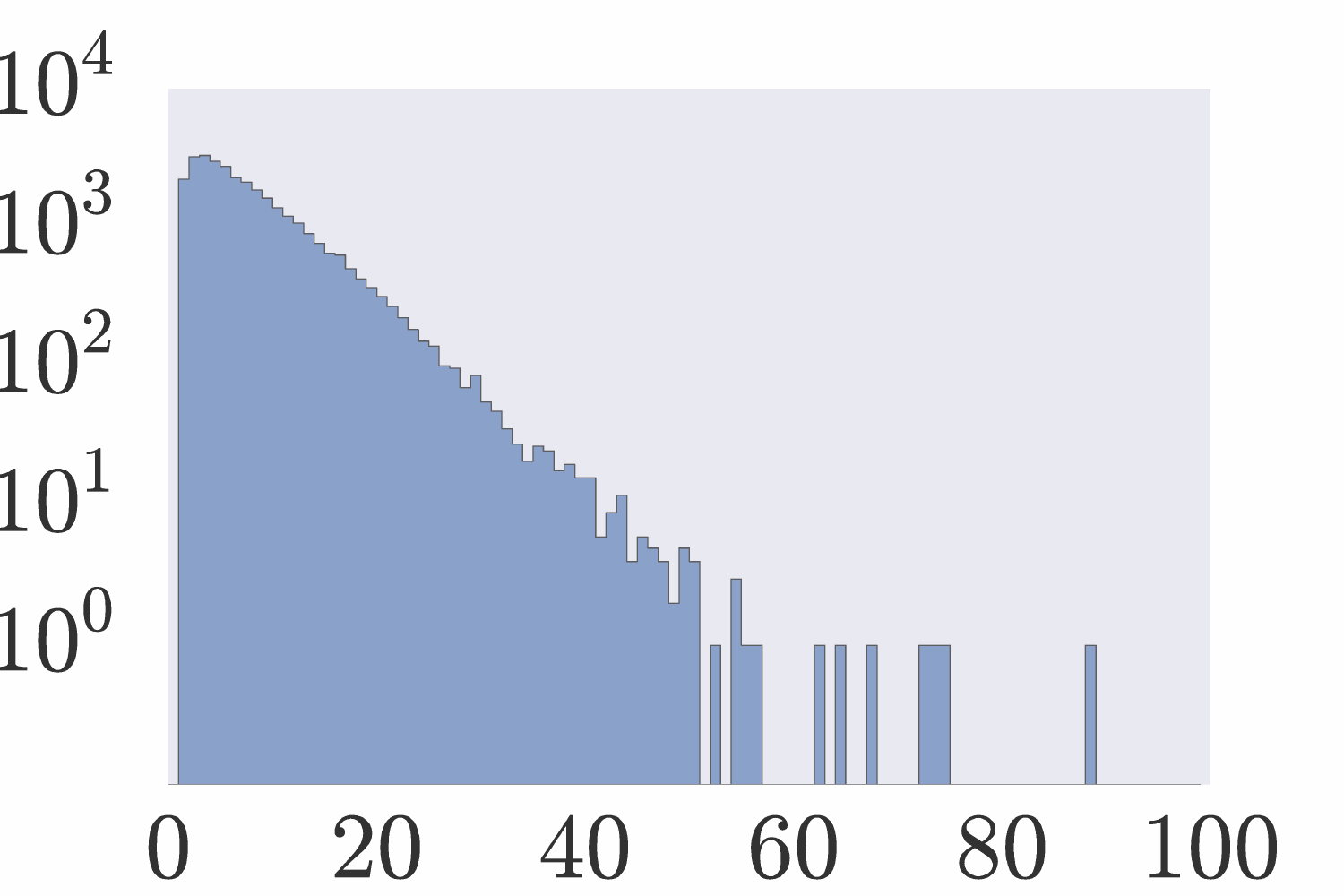}
\caption{\small Dependencies.}
\end{subfigure}
\caption{\small Histograms of statement lengths, occurrences of each word, and statement
dependencies in the Mizar corpus translated to first order logic.  The wide length distribution
poses difficulties for RNN models and batching, and many rarely occurring words
make it important to take definitions of words into account. \label{stats}}
\end{figure}

The Mizar data set is also an interesting case study in neural network
sequence tasks, as it differs from natural language problems in several ways.
It is highly structured with a simple context free grammar -- the interesting
task occurs only after parsing.  The distribution of lengths is
wide, ranging from 5 to 84,299 characters with mean 304.5, and from
2 to 21,251 tokens with mean 107.4 (see Figure~\ref{stats}).  Fully recurrent
models would have to back-propagate through 100s to 1000s of characters or 100s
of tokens to embed a whole statement.  Finally, there are many rare words
-- 60.3\% of the words occur fewer
than 10 times -- motivating the definition-aware embeddings in
section~\ref{word-level}.

\section{Overview of our approach}
\begin{figure}
\centering
\begin{subfigure}{.35\textwidth}
\includegraphics[width=\linewidth]{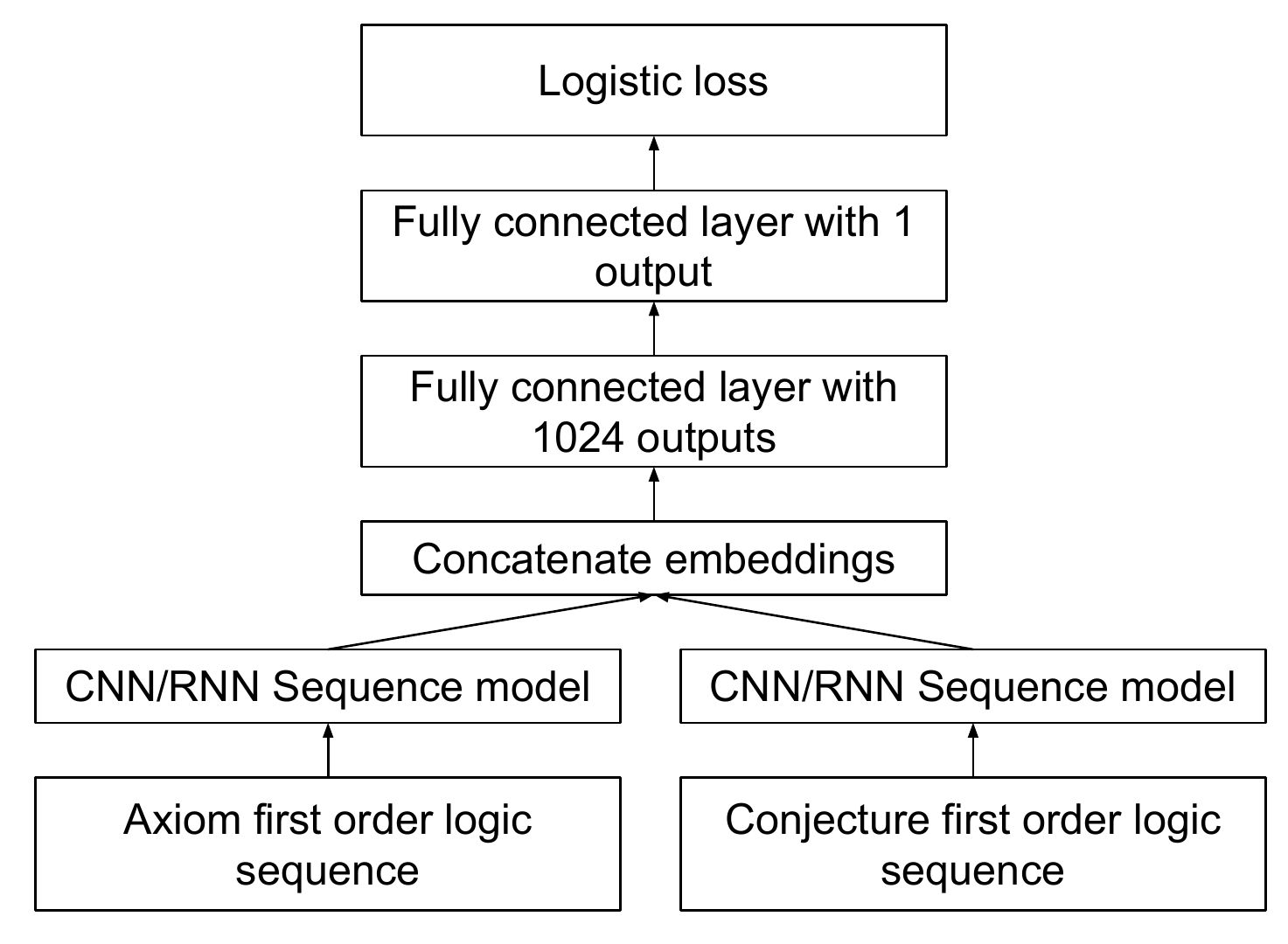}
\end{subfigure}
\hspace{0.1\textwidth}
\begin{subfigure}{.45\textwidth}
 \includegraphics[width=\linewidth]{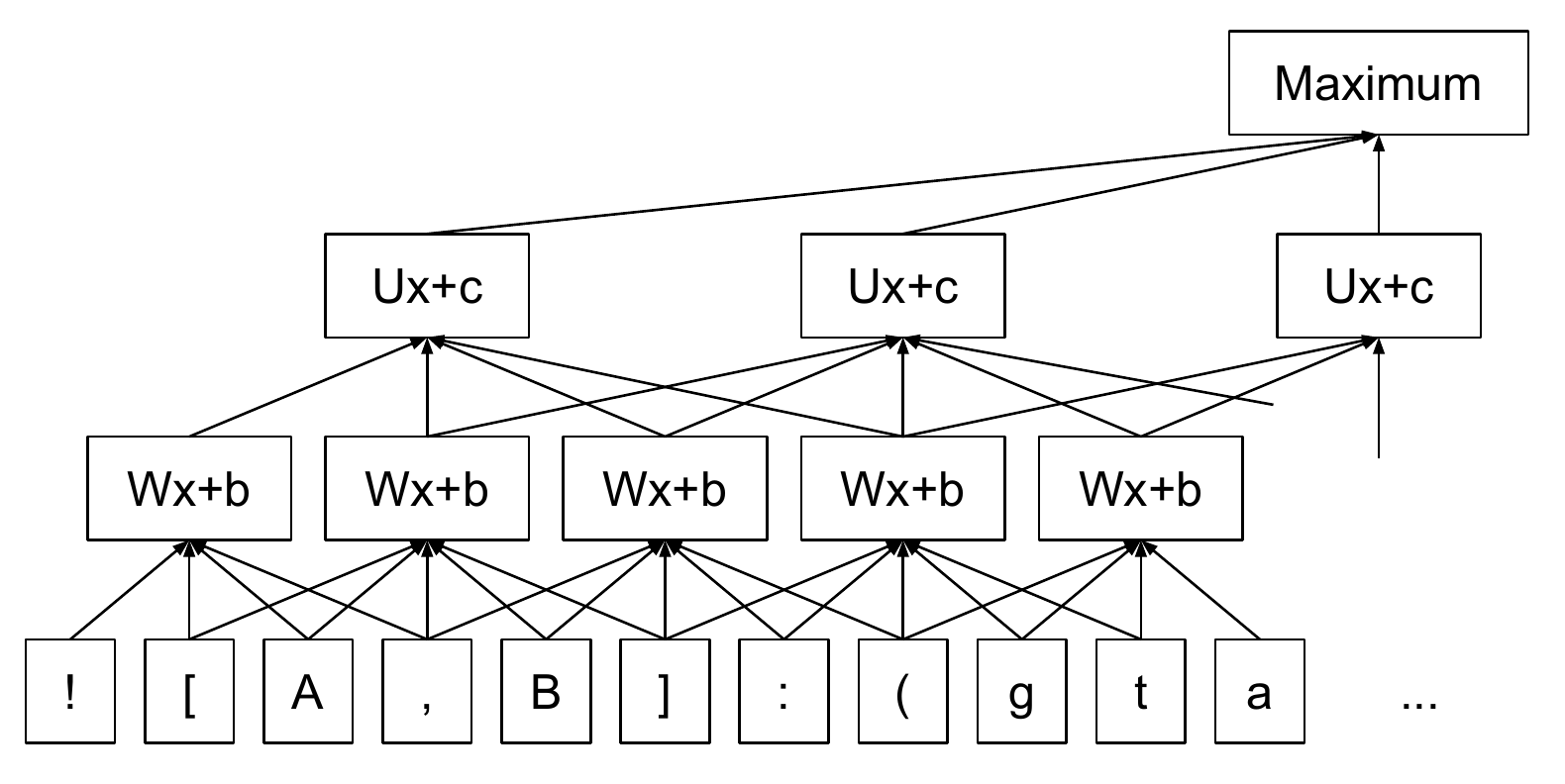}
\end{subfigure}
\caption{\small (left) Our network structure.
  The input sequences are either character-level (section~\ref{char-level}) or
  word-level (section~\ref{word-level}). We use separate models to embed
  conjecture and axiom, and a logistic layer to predict
  whether the axiom is useful for proving the conjecture.
  (right) A convolutional model.
  \label{fig:overall}
}
\vspace{-.2in}
\end{figure}

The full premise selection task takes a conjecture and a set of axioms and
chooses a subset of axioms to pass to the ATP.  We simplify
from subset selection to pairwise relevance by predicting the
probability that a given axiom is useful for proving a given conjecture.
This approach depends on a relatively sparse dependency
graph. 
Our general
architecture is shown in Figure~\ref{fig:overall}(left): the conjecture and axiom sequences
are separately embedded into fixed length real vectors, then concatenated and
passed to a third network with two fully connected layers and logistic loss.
During training time, the two embedding networks and the joined
predictor path are trained jointly.

As discussed in section~\ref{premise}, we train our models on premise selection
data generated by a combination of various methods, including k-nearest-neighbor search on
hand-engineered similarity metrics.
We start with a first stage of character-level models, and then build second and later
stages of word-level models on top of the results of earlier stages.

\subsection{Stage 1: Character-level models \label{char-level}}
We begin by avoiding special purpose engineering by treating formulas on the
character-level using an 80 dimensional one-hot encoding of the character sequence.
These sequences are passed to a weight shared network for variable length input.
For the embedding computation, we have explored the following architectures:
\begin{small}
\begin{enumerate}
  \item Pure recurrent LSTM~\cite{hochreiter1997long} and GRU~\cite{chung2015gated}
    networks.
  \item A pure multi-layer convolutional network with various numbers of
	  convolutional layers (with strides) followed by a global temporal max-pooling reduction (see Figure~\ref{fig:overall}(right)).
  \item A recurrent-convolutional network, that uses convolutional layers to produce
        a shorter sequence which is processed by a LSTM.
\end{enumerate}
\end{small}

The exact architectures used are specified in the experimental section.

It is computationally prohibitive to compute a large number of
(conjecture, axiom) pairs due to the costly embedding phase. Fortunately,
our architecture allows caching the embeddings for conjectures and axioms
and evaluating the shared portion of the network for a given pair. This makes it
practical to consider all pairs during evaluation.

\subsection{Stage 2: Word-level models \label{word-level}}
The character-level models are limited to word and structure similarity within
the axiom or conjecture being embedded.  However, many of the symbols occurring
in a formula are defined by formulas earlier in the corpus, and we can use the
axiom-embeddings of those symbols to improve model performance.

Since Mizar is based on first-order set theory, definitions of symbols can be
either explicit or implicit.  An explicit definition of $x$ sets $x = e$ for
some expression $e$, while an implicit definition states a property of the
defined object, such as defining a function $f(x)$ by
$\forall x. f(f(x)) = g(x)$.
To avoid manually encoding the structure of implicit definitions, we
embed the entire statement defining a symbol $f$, and then use the
stage 1 axiom-embedding corresponding to the whole statement as a word-level
embeddings.

Ideally, we would train a single network that embeds statements by recursively
expanding and embedding the definitions of the defined symbols.  Unfortunately, this recursion
would dramatically increase the cost of training since the definition chains
can be quite deep.  For example, Mizar defines real numbers in
terms of non-negative reals, which are defined as Dedekind cuts of
non-negative rationals, which are defined as ratios of naturals, etc.
As an inexpensive alternative, we reuse the axiom
embeddings computed by a previously trained character-level model, mapping each
defined symbol to the axiom embedding of its defining
statement.  Other tokens such as brackets and
operators are mapped to fixed pseudo-random vectors of the same dimension.

Since we embed one token at a time ignoring the grammatical structure, our approach
does not require a parser: a trivial lexer is implemented in a few lines of
Python. With word-level embeddings, we use the same architectures with shorter
input sequence to produce axiom and conjecture embeddings for ranking
the (conjecture, axiom) pairs.
Iterating this approach by using the resulting, stronger axiom embeddings
as word embeddings multiple times for additional stages did not yield
measurable gains.

\section{Experiments}

\subsection{Experimental Setup}

For training and evaluation we use a subset of 32,524 out of 57,917 theorems
that are known to be provable by an ATP given the right set of premises.
We split off a random 10\% of these (3,124 statements) for testing and validation.
Also, we held out 400 statements from the 3,124 for monitoring
training progress, as well as for model and checkpoint selection.
Final evaluation was done on the remaining
2,724 conjectures. Note that we only held out conjectures, but we trained on all statements as axioms.
This is comparable to our k-NN baseline which is also trained on all statements as axioms. 
The randomized selection of the training and testing sets may also lead to learning
from future proofs: a proof $P_j$ of theorem $T_j$ written after theorem $T_i$
may guide the premise selection for $T_i$.
However, previous $k$-NN experiments show similar performance between a full
10-fold cross-validation and incremental evaluation as long as chronologically
preceding formulas participate in proofs of only later theorems.

\subsection{Metrics}

For each conjecture, our models output a ranking of possible premises.
Our primary metric is the number of conjectures proved from
the top-$k$ premises, where $k = 16, 32, \ldots, 1024$.
This metric can accommodate alternative proofs but is computationally
expensive. Therefore we additionally measure the ranking quality using
the average maximum relative rank of the testing premise set.
Formally, average max relative rank is
\begin{align*}
\textrm{aMRR} = \underset{C}{\operatorname{mean}}
  \underset{P \in \mathcal{P}_\textrm{test}(C)}{\max}
  \frac{\operatorname{rank}(P, \mathcal{P}_\textrm{avail}(C))}{|\mathcal{P}_\textrm{avail}(C)|}
\end{align*}
where $C$ ranges over conjectures, $\mathcal{P}_\textrm{avail}(C)$ is the set of
premises available to prove $C$, $\mathcal{P}_\textrm{test}(C)$ is the set of premises for conjecture
$C$ from the test set, and $\operatorname{rank}(P, \mathcal{P}_\textrm{avail}(C))$
is the rank of premise $P$ among the set $\mathcal{P}_\textrm{avail}(C)$ according to the model.
The motivation for aMRR is that conjectures are easier to prove if all their dependencies occur
early in the ranking.

Since it is too expensive to rank all
axioms for a conjecture during continuous evaluation, we approximate
our objective.  For our holdout set of 400 conjectures,
we select all true dependencies $\mathcal{P}_\textrm{test}(C)$
and 128 fixed random false dependencies from
$\mathcal{P}_\textrm{avail}(C) - \mathcal{P}_\textrm{test}(C)$
and compute the average max relative rank in this ordering.
Note that aMRR is nonzero even if all true dependencies are ordered before
false dependencies; the best possible value is 0.051.

\subsection{Network Architectures}

\begin{figure}
\centering
\includegraphics[width=1.\linewidth]{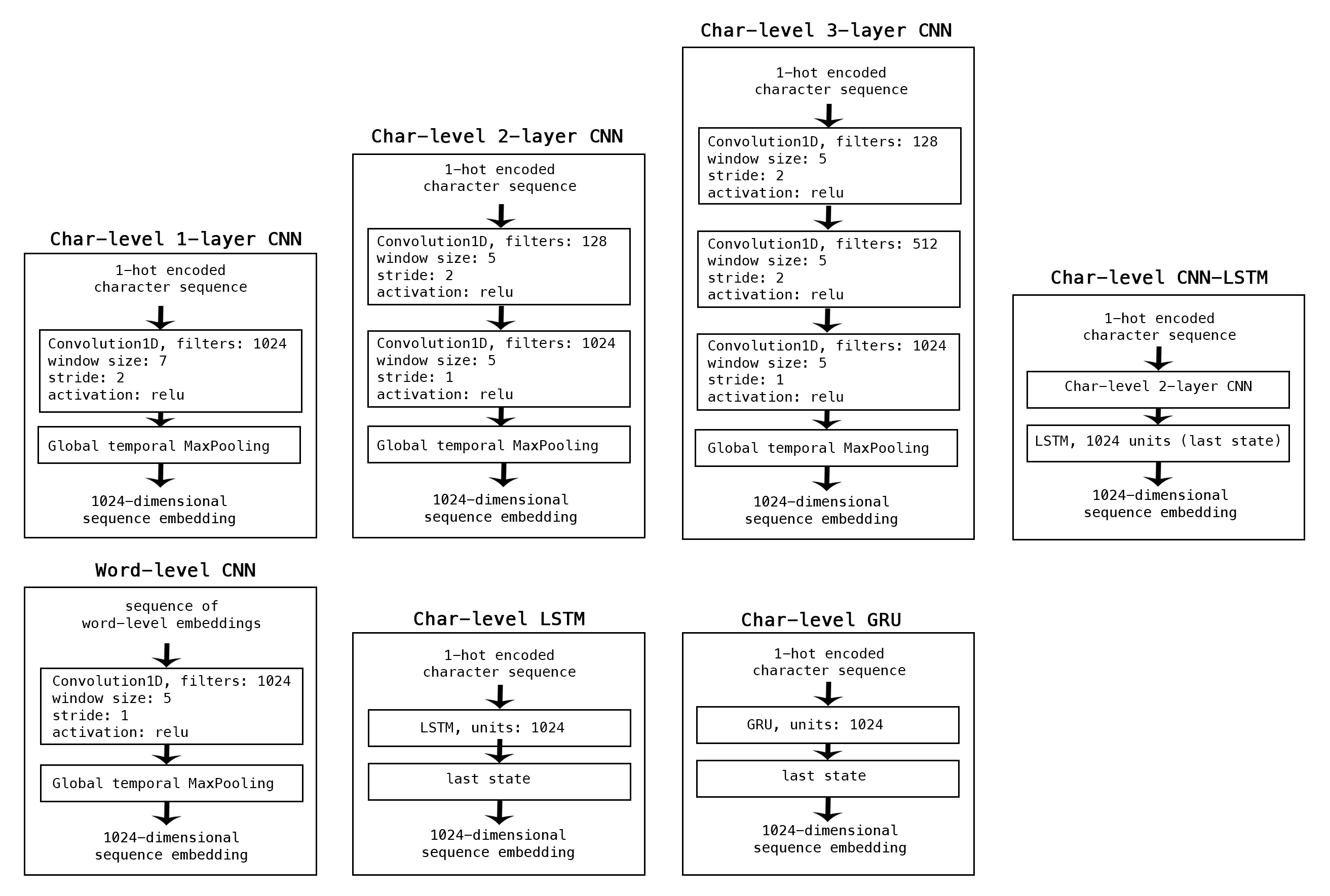}
\caption{Specification of the different embedder networks.}
\label{fig:networks}
\end{figure}

All our neural network models use the general architecture from
Fig~\ref{fig:overall}: a classifier on top of the concatenated embeddings of an
axiom and a conjecture.  The same classifier architecture was used for all
models: a fully-connected neural network with one hidden layer of size 1024.
For each model, the axiom and conjecture embedding networks have the same
architecture without sharing weights. The details of the embedding networks are
shown in Fig~\ref{fig:networks}.  

\subsection{Network Training}

The neural networks were trained using asynchronous distributed stochastic
gradient descent using the Adam optimizer~\cite{kingma2014adam} with up to 20
parallel NVIDIA K-80 GPU workers per model.  We used the TensorFlow
framework~\cite{tensorflow2015-whitepaper} and the Keras
library~\cite{chollet2015keras}. The weights were initialized using
\cite{glorot2010understanding}. Polyak averaging with 0.9999 decay
was used for producing the evaluation weights~\cite{polyak1992acceleration}.
The character level models were trained with maximum sequence length 2048
characters, where the word-level (and definition embedding) based models had
a maximum sequence length of 500 words. For good performance,
especially for low cutoff thresholds, it was critical to employ negative
mining during training. A side process was continuously
evaluating many (conjecture, axiom) pairs. For each conjecture,
we pick the lowest scoring statements that have higher score than
the lowest scoring true positive. A queue of previously mined negatives is
maintained for producing a mixture of examples in which the ratio of mined
instances is about 25\% and the rest are randomly selected premises.
Negative mining was crucial for good quality: at the top-16 cutoff,
the number of proved theorems on the test set has doubled. For the union of
proof attempts over all cutoff thresholds, the ratio of successful proofs
has increased from 61.3\% to 66.4\% for the best neural model.

\subsection{Experimental Results}
Our best selection pipeline uses a stage-1 character-level
convolutional neural network model to produce word-level embeddings for the
second stage. The baseline uses distance-weighted
$k$-NN~\cite{EasyChair:74,KaliszykU13b} with
handcrafted semantic features~\cite{KaliszykUV15a}.
For all conjectures in our holdout set, we consider all
the chronologically preceding statements (lemmas, definitions and axioms) 
as premise candidates. In the DeepMath case, premises
were ordered by their logistic scores. E prover was applied to
the top-$k$ of the premise-candidates for each of the cutoffs
$k\in(16, 32, \ldots, 1024)$ until a proof is found or $k = 1024$ fails.
Table~\ref{tab:results} reports the number of theorems proved
with a cutoff value {\it at most} the $k$ in the leftmost column.
For E prover, we used auto strategy with a soft time limit of 90 seconds,
a hard time limit of 120 seconds, a memory limit of 4 GB,
and a processed clauses limit of 500,000.

Our most successful models employ simple convolutional networks followed by
max pooling (as opposed to recurrent networks like LSTM/GRU), and the
two stage definition-based {\bf def-CNN} outperforms the na\"{\i}ve
{\bf word-CNN} word embedding significantly. In the latter the word embeddings
were learned in a single pass; in the former they are fixed
from the stage-1 character-level model. For each architecture (cf.
Figure~\ref{fig:networks}) two convolutional layers perform
best. Although our models differ significantly from each other, they differ even
more from the $k$-NN baseline based on hand-crafted features. The right column
of Table~\ref{tab:results} shows the result if we average the prediction score of the stage-1
model with that of the definition based stage-2 model. We also experimented
with character-based RNN models using shorter sequences:
these lagged behind our long-sequence CNN models but performed significantly
better than those RNNs trained on longer sequences. This suggest that
RNNs could be improved by more sophisticated optimization techniques such as
curriculum learning.


\begin{figure}
\centering
\begin{subfigure}{.47\textwidth}
\includegraphics[width=.9\linewidth]{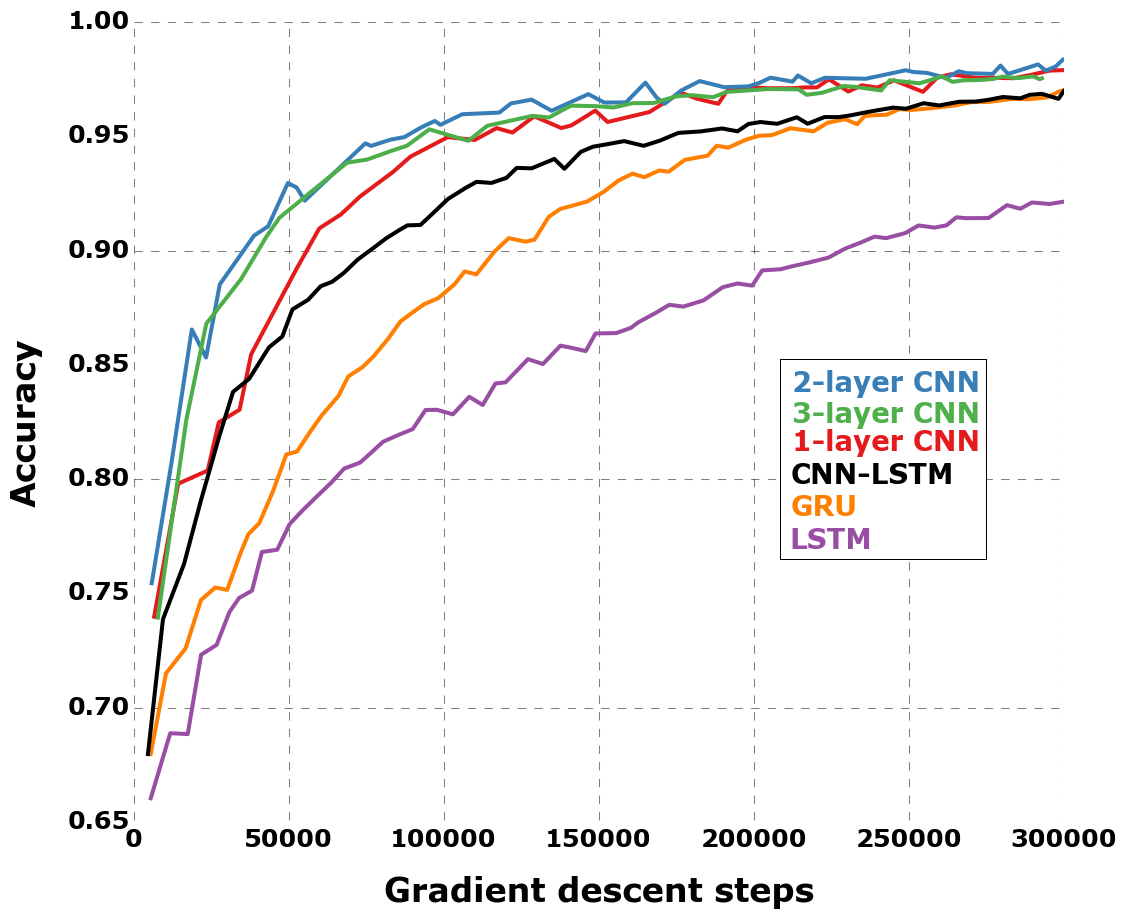}
\caption{\small Training accuracy for different character-level models without hard negative mining. Recurrent models seem underperform, while pure convolutional models yield the best results. For each architecture, we trained three models with different random initialization seeds. Only the best runs are shown on this graph; we did not see much variance between runs on the same architecture.}
\label{fig:nips_acc_per_architecture}
\end{subfigure}\hspace{0.03\textwidth}
\begin{subfigure}{.47\textwidth}
\includegraphics[width=.9\linewidth]{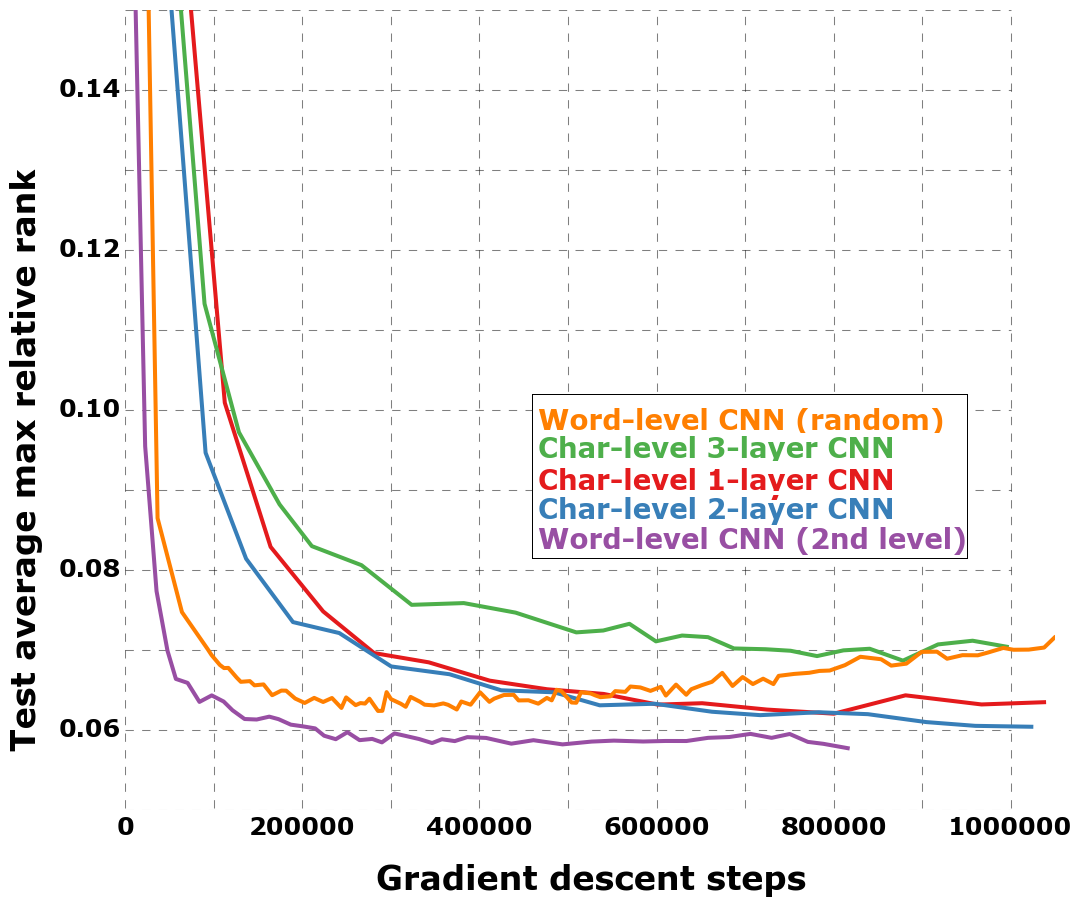}
\caption{\small Test average max relative rank for different models without hard negative mining. The best is a word-level CNN using definition embeddings from a character-level 2-layer CNN. An identical word-embedding model with random starting embedding overfits after only 250,000 iterations and underperforms the best character-level model.}
\label{fig:nips_avmraxrank_per_architecture}
\end{subfigure}
\end{figure}
\begin{table}[htbp]
\begin{center}
	\resizebox{\textwidth}{!}{%
		\begin{tabular}[H]{r|c|c|c|c|c|c}
	   {\bf Cutoff} & {\bf $k$-NN Baseline} (\%) & {\bf char-CNN} (\%) & {\bf word-CNN} (\%) & {\bf def-CNN-LSTM} (\%) & {\bf def-CNN} (\%) & {\bf def+char-CNN} (\%)  \\
           \hline
           16 &         674     (24.6)& 687     (25.1)   &    709  (25.9)       &       644  (23.5)    &       734  (26.8)    & \bf   835  (30.5)    \\ \hline
           32 &         1081    (39.4)& 1028    (37.5) &    1063   (38.8)    &       924    (33.7)  &       1093   (39.9) & \bf   1218    (44.4) \\ \hline
           64 &         1399    (51)   & 1295 (47.2)    &    1355  (49.4)     &       1196  (43.6)  &       1381   (50.4) & \bf   1470    (53.6) \\ \hline
           128 &        1612    (58.8)& 1534    (55.9)&    1552    (56.6)   &       1401    (51.1)&       1617    (59)  & \bf   1695     (61.8)\\ \hline
           256 &        1709    (62.3)& 1656    (60.4)&    1635    (59.6)   &       1519    (55.4)&       1708    (62.3)& \bf   1780     (64.9)\\ \hline
           512 &        1762    (64.3)& 1711    (62.4)&    1712    (62.4)   &       1593    (58.1)&       1780    (64.9)& \bf   1830     (66.7)\\ \hline
           1024 &      1786   (65.1) & 1762    (64.3)&    1755     (64)    &       1647    (60.1)&       1822    (66.4)&\bf   1862     (67.9)\\ \hline 
	\end{tabular}}
\end{center}
  \caption{\small Results of ATP premise selection experiments with hard negative mining on a test set of 2,742 theorems. 
        Each entry is the number (\%) of theorems
	proved by E prover using that particular model to rank the premises.
        The union of def-CNN and char-CNN proves 69.8\% of the test set, while the union of the
        def-CNN and k-NN proves 74.25\%. This means that the neural network predictions
        are more complementary to the k-NN predictions than to other neural models.
The union of all methods proves 2218 theorems (80.9\%) and just the neural models prove 2151 (78.4\%).}
  \label{tab:results}
\end{table}

\begin{figure}
\centering
\begin{subfigure}{.47\textwidth}
\includegraphics[width=.6\linewidth]{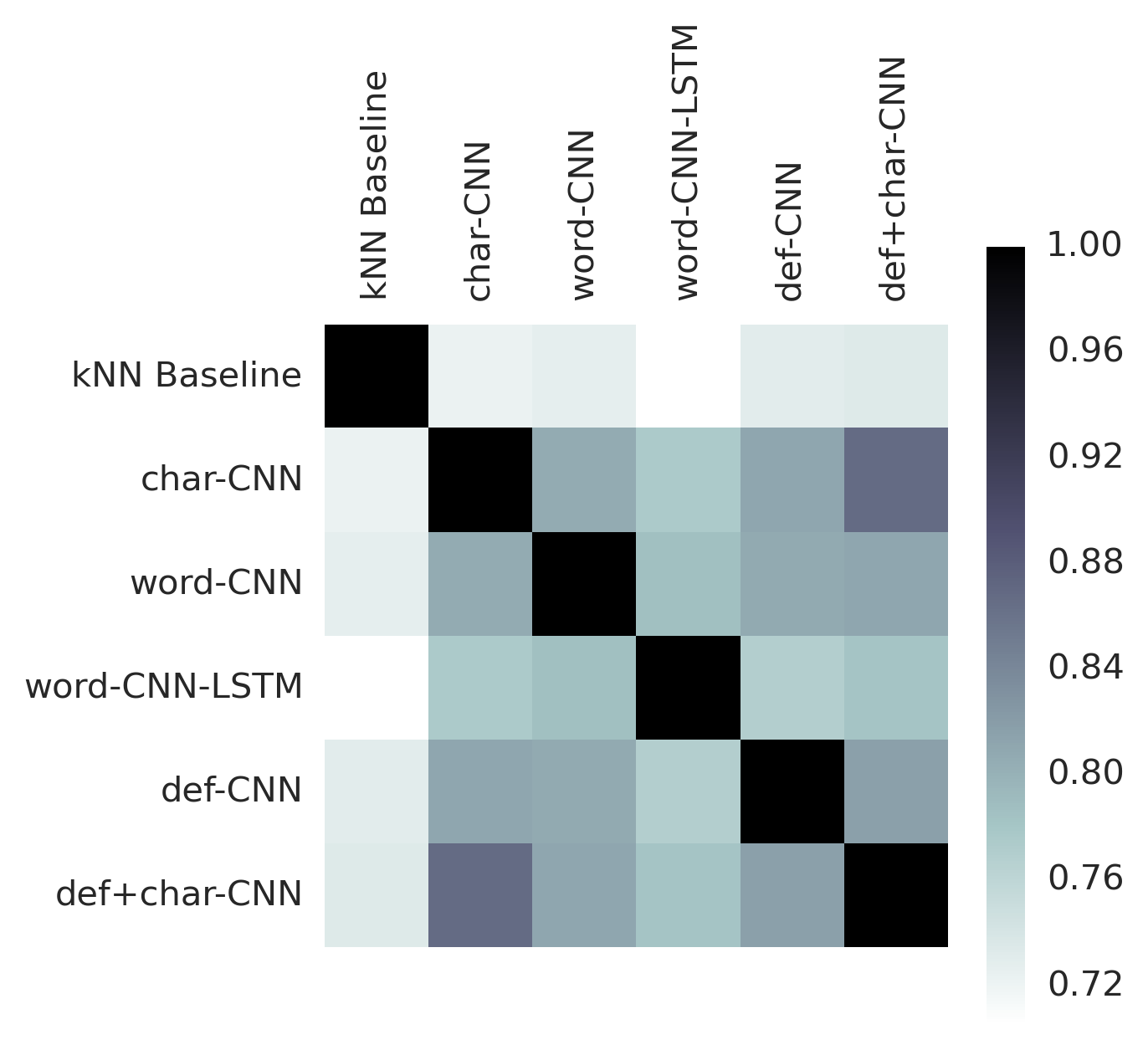}
	\caption{\small Jaccard similarities between proved sets of conjectures across models. Each of the neural network
	model prediction are more like each other than those of the $k$-NN baseline.}
\label{fig:jaccard_sims}
\end{subfigure}\hspace{0.03\textwidth}
\begin{subfigure}{.47\textwidth}

    \resizebox{\textwidth}{!}{%
    \begin{tabular}[H]{r|c|c|c|c|c|c}
     {\bf Model} & {\bf Test min average relative rank}   \\
   \hline
     char-CNN &   0.0585  \\ \hline
     word-CNN &   {0.06}  \\ \hline
     def-CNN-LSTM & {0.0605} \\ \hline
     def-CNN & \bf 0.0575 \\ \hline
    \end{tabular}}
    \caption{\small Best sustained test results obtained by the above models.
             Lower values are better. This was monitored continuously during
             training on a holdout set with 400 theorems, using all true
             positive premises and 128 randomly selected negatives.
             In this setup, the lowest attainable max average relative rank
             with perfect predictions is $0.051$.
    }

\end{subfigure}
\end{figure}

Also, when we applied two of the premise selection models on those Mizar
statements that were not proven automatically before, we managed to prove
823 additional of them.

\section{Conclusions}
In this work we provide evidence that even simple
neural models can compete with hand-engineered features for premise selection,
helping to find many new proofs. This translates to real gains in automatic
theorem proving. Despite these encouraging results, our models are relatively
shallow networks with inherent limitations to representational power and are
incapable of capturing high level properties of mathematical statements.
We believe theorem proving is a challenging and important domain for deep
learning methods, and that more sophisticated optimization techniques and
training methodologies will prove more useful than in less structured domains.

\section{Acknowledgments}
We would like to thank Cezary Kaliszyk for providing us with an improved
baseline model. Also many thanks go to the Google Brain team for their generous
help with the training infrastructure. We would like to thank Quoc Le for
useful discussions on the topic and to Sergio Guadarrama for his help with
TensorFlow-slim.